\RequirePackage{fix-cm}

\documentclass[twocolumn]{svjour3}         

\smartqed  

\usepackage{graphicx}
\usepackage{tabularx}
\usepackage{adjustbox}
\usepackage{multicol}
\usepackage{array}
\usepackage{multirow}
\usepackage{booktabs}

\begin{document}

\title{Robots in healthcare as envisioned by care professionals}

\titlerunning{}    

\author{Fran Soljacic         \and
        Meia Chita-Tegmark    \and
        Theresa Law           \and
        Matthias Scheutz
}

\authorrunning{Soljacic et al.} 

\institute{F. Soljacic \at
              Dept. of Community Health \\
              Tufts University\\
              Medford, MA\\
              \email{fran.soljacic@tufts.edu}        
           \and
           M. Chita-Tegmark \at
              Dept. of Computer Science \\
              Tufts University\\
              Medford, MA\\
              \email{mihaela.chita\_tegmark@tufts.edu}\\
              \and
          T. Law \at              
              Dept. of Computer Science \\
              Tufts University\\
              Medford, MA\\
              \email{theresa.law@tufts.edu}\\
              \and
        M. Scheutz \at
              Dept. of Computer Science \\
              Tufts University\\
              Medford, MA\\
              \email{matthias.scheutz@tufts.edu}
}

\date{Received: date / Accepted: date}

\maketitle
\sloppy

\begin{abstract}
As AI-enabled robots enter the realm of healthcare and caregiving, it is important to consider how they will address the dimensions of care and how they will interact not just with the direct receivers of assistance, but also with those who provide it (e.g., caregivers, healthcare providers etc.). Caregiving in its best form addresses challenges in a multitude of dimensions of a person's life: from physical, to social-emotional and sometimes even existential dimensions (such as issues surrounding life and death). In this study we use semi-structured qualitative interviews administered to healthcare professions with multidisciplinary backgrounds (physicians, public health professionals, social workers, and chaplains) to understand their expectations regarding the possible roles robots may play in the healthcare ecosystem in the future. We found that participants drew inspiration in their mental models of robots from both works of science fiction but also from existing commercial robots. Participants envisioned roles for robots in the full spectrum of care, from physical to social-emotional and even existential-spiritual dimensions, but also pointed out numerous limitations that robots have in being able to provide comprehensive humanistic care. While no dimension of care was deemed as exclusively the realm of humans, participants stressed the importance of caregiving humans as the primary providers of comprehensive care, with robots assisting with more narrowly focused tasks. Throughout the paper we point out the encouraging confluence of ideas between the expectations of healthcare providers and research trends in the human-robot interaction (HRI) literature. 
\end{abstract}


\section{Introduction}

As progress in  artificial intelligence (AI) in the past decade has enabled the development of robots capable of ever-more sophisticated tasks, visions for using assistive robots in healthcare and caregiving have also grown in ambition. AI-enabled robots are being developed to fulfill numerous purposes in healthcare. The robotic system PHAROS, for example, uses deep learning to analyze users' exercise capabilities and create a personal exercise routine \cite{fiorini2019assistive,martinez2020socially,kyrarini2021survey}. Another example is the Hobbit live-in assistive robot that is capable of autonomously navigating around a patient's home, emergency detection and response, retrieving objects for the patient, and entertainment \cite{fischinger2016hobbit}. Many other AI-enabled robots have been introduced for general mobility assistance, feeding assistance, transferring a patient between a wheelchair and their bed, monitoring, bathing, and providing emotional and cognitive care \cite{maalouf2018robotics}. Moreover, the COVID-19 pandemic has brought to the forefront risks specific to infectious diseases coming from person-to-person contact, thus shedding new light on the numerous roles that robots could play in disinfecting \cite{guettari2020uvc,conte2020design}, mitigating the negative health effects of social isolation \cite{jecker2020you}, and even enforcing social distancing \cite{murphy2020applications}.

As AI-enabled robots enter the realm of healthcare and caregiving it is important to consider how they will address the many different facets of care and how they will interact not just with the direct receivers of assistance but also with those who provide it (e.g., caregivers, healthcare providers, etc.). Caregiving in its best form addresses challenges in a multitude of dimensions of a person's life: from physical, to social-emotional and sometimes even existential dimensions, such as issues surrounding life and death. Health outcomes have been shown to be interlinked with many factors outside of the purely physical realm, such as having social support \cite{kaplan1977social,schwarzer1991social}, feeling hopeful \cite{hinds1988hopefulness,mok2010health}, and finding meaning in life \cite{candela2020finding,bower1998cognitive}. In response to this multidimensional set of needs, many healthcare ecosystems provide access in addition to doctors and nurses to occupational therapists, social workers and even chaplains who, through their services, extend the dimensionality of compassionate care \cite{lefevor2021religiousness}. Understanding where robots fit in this ecosystem and how they best can help complement humans in meeting not only the physical but also the social, cultural, and deeply personal dimensions of care is paramount to ensure that robots will not be disruptive and that their presence will not lead to a neglectful approach to deeper aspects of care.

This paper seeks to understand the expectation of healthcare professionals: doctors, social workers, public health professors, hospital and university chaplains, and other healthcare professionals regarding the introduction of robots in the healthcare ecosystem. We conducted interviews with different healthcare professionals focusing on the specific roles they saw robots playing in the healthcare and caregiving ecosystem, in both private homes as well as in hospitals and other care facilities, in more typical circumstances as well as during the COVID pandemic, the largest public health crisis in recent decades. We specifically pursued the following research question: how do various healthcare professionals view robots in terms of their abilities, and what roles do they envision for them in healthcare and caregiving? In our interviews we also touched on healthcare professionals' general knowledge of, as well as thoughts and attitudes towards, robots.  An overarching theme we explore in this paper is the potential of robots to provide ``humanistic care,'' where patients are cared for not just physically but also emotionally and socially in seemingly empathetic and thoughtful ways.  

What is novel about our study is that it integrates perspectives from a variety of healthcare professions and it explores the multidimensionality of caregiving including the social-emotional and spiritual-existential dimensions in addition to the physical ones. This exploration makes several contributions to the HRI field: (1) it offers ideas for robot applications that originate from professionals intimately familiar with different aspects of caregiving, (2) it gives a snapshot of how robots are perceived from the point of view of care providers, (3) it highlights concerns about robotic caregiving that need to be mitigated in order for robots to be a successful addition to the care ecosystem, and finally, (4) it gives a critical analysis of the robots potential to fulfill or complement the deeper and more subtle aspects of humanistic care.

The paper is organized as follows: the background section offers an overview of proposed roles for robots in the healthcare ecosystem in the HRI literature. In the methods section we present the interview procedures, the participants in our study, as well as the analytical approach we undertook in engaging with the data collected through our interviews. Finally, we explore in depth the themes identified in the interviews and connect them to the HRI literature and other participatory-science studies conducted with different groups. We discuss general attitudes of healthcare professionals towards robots, the articulation of specific visions for tasks and roles robots could undertake in the healthcare ecosystem as imagined by various healthcare professionals, and finally concerns and perceived limitations of robots and their potential for humanistic care. We conclude by discussing the implications of these findings for HRI.

\section{Background}

The role of caregiver goes beyond what trained healthcare professionals offer. A survey found that 25.3\% of Americans act as informal, unpaid caregivers for a loved one or family member \cite{trivedi2014characteristics}. This already heavy burden will only worsen as the population of developed countries ages due to decreasing birth rates.  Japan's official health policy incorporates widespread use of assistive robots to help with its aging population \cite{sharkey2012granny}. While most commercial assistive robots up until now have focused on helping with tasks such as surgery \cite{davies2000review}, other advanced AI-enabled robots that can help with multiple tasks have been designed and produced \cite{bedaf2015overview}. As technology advances and manufacturing costs go down, it is likely that assistive robots will be able to take over many of the tasks currently performed by informal caregivers \cite{van2013designing}. These robots will have to operate semi-autonomously, allowing for technicians and even family members to take control of the robot in case of issues such as patient confusion \cite{mast2012user}. Additionally, the COVID-19 pandemic has shown that robots also have many potential roles that they can fill to help improve people's lives while limiting the spread of infectious disease. These roles include acting as telepresence devices to limit in-person gatherings, encouraging physical activity during times of limited mobility, and supporting mental health during periods of social isolation \cite{scassellati2020potential}.

 Socially assistive robots provide assistance to people through social rather than physical interactions \cite{feil2005defining}. Some socially assistive robots (SARs) are already commercially available and have been shown to have positive effects. For example, after using the robot Paro, a pet like companion robot which is designed to look and behave like a baby seal, nursing home patients were more active and social with those around them \cite{vsabanovic2013paro}. SARs have already been used for education, weight loss, autism spectrum disorder, dementia, and stroke rehabilitation \cite{rabbitt2015integrating}. 

In addition to directly helping patients themselves, a robot could also help a caregiver by reducing the emotional burden that can come with providing care to someone. The robot can become an authoritative figure by communicating information that the patient does not want to hear, providing redirection during emotionally difficult times, and facilitating interactions between caregiver and patient \cite{moharana2019robots}. Additionally, as the physical technological design of robots progresses, they will increasingly be able to take over physically demanding, or monotonous tasks or tasks requiring privacy (e.g., bathing) from human caregivers. When asked what tasks a robot could help with the most, elderly people and caregivers ranked emergency detection and response, reaching objects, and various household chores as the most useful \cite{mast2012user}. Robots are also being designed to assist a patient with intimate tasks such as using the restroom and bathing; the goal of these robots in to help patients preserve their sense of independence and dignity \cite{van2013designing}.Whether a robot is performing physically assistive tasks or social/emotionally assistive tasks for a patient, there is concern that robotic care would be inherently inferior to human care because today's robots are incapable of exhibiting genuine empathy and compassion \cite{parks2010lifting,sparrow2006hands}. Coghlan introduces the idea of ``humanistic care'' as a phrase for these seemingly uniquely human aspects of caregiving \cite{coghlan2021robots}. Several studies have explored how to design robots to emulate human touch, compassion, and empathy. Block and Kuchenbecker built PR2, a robot designed to emulate human hugs and then measured participants' responses to factors such as hug temperature, pressure, and duration \cite{block2019softness}. Niculescu et al. experimented with robot's personalities, voices, and humor to determine how to design the most pleasant socially assistive robots. James et al. created a robotic voice that mimicked empathy and found that participants recognized it as empathetic and preferred it to a non-empathetic robot voice \cite{james2018artificial}.

Multiple studies exploring people's models and ideas for future uses of robotics have used qualitative analysis of interviews with a small number of participants ($n<20$). Horstmann et al. \cite{horstmann2019great} used an online survey to determine lay-people's perceptions, concerns, and expectations of social robotics. This was accompanied by a smaller study of 13 semi-structured qualitative interviews. This methodology allows for the analysis of a complex topic even with a relatively low number of participants. The study found that participants' had high expectations of the abilities of robots and their mental models were largely influenced by media depictions of robots. Walden et al. \cite{walden2015mental} conducted qualitative interviews of 45 senior citizens to determine their preferences for physically and socially assistive robots. They found that seniors' mental models of robots were highly influenced by representations of robots in media and popular culture. In Switzerland, Ray et al.  \cite{ray2008people} conducted over 240 surveys and 11 interviews to assess participant's positive and negative images of robotics and to identify which tasks a lay person can imagine being performed by a robot in the near future. O'Leary et al. \cite{o2020community} used participatory design methodology for understanding how to culturally tailor virtual agents that delivered health promotion interventions. Participants were members of predominantly black communities, who provided input on what aspects of their lives they'd like help with, how the agents should look like and the functionality it should have, as well as tailing health conversations with the agent to include content that was relevant on both a personal and community level by, for example, including religious and scriptural references. Other qualitative studies include Moharana et al.'s interviews with caregivers of people with dementia \cite{moharana2019robots} and Williams et al.'s qualitative evaluation of the design of the AIDA robot through observation of users' interactions \cite{williams2019aida}.

\section {Methods}

Our study is inspired by previous qualitative studies using semi-structured interviews for in-depth exploration of topics in robotic caregiving. What is novel about this study is that we interviewed participants with different professional backgrounds capable of providing expert opinions on multiple dimensions of caregiving, from physical to social-emotional and even spiritual-existential.

\subsection{Procedure}

Interviews were conducted over Zoom and lasted about 1-hour. Prior to the interview participants completed a short survey to capture demographic information and participants' mental models of robots (e.g., what robots look like and what they are able to do). One of the questions in this survey listed 26 activities and asked the participants to check off each activity that they believed a robot would be able to help with. During the interview, the interviewer would select three activities that the participant checked off in the survey and then ask participants to elaborate on how a robot would help with each. 

The interviews were audio recorded, transcribed using Google transcription software and inspected for correctness. The structured interview was scripted with slight modifications depending on the participant's profession. The focus of this paper is the initial portion of the interview which consisted of questions designed to further capture information about participants' mental models of robots in addition to questions asking participants to imagine roles for robots in caregiving.

The interview questions included in this analysis along with explanations as to why each question was chosen are available in the supplementary materials. A complete copy of the interview questions, including those that were not included in this paper's analysis, is also available in the supplementary materials. 

The questions that are the focus of this paper asked about participants' mental models of robots, how they imagine a robot providing help in daily life or social/emotional matters, and if they could see themselves or a loved one being cared for by a robot. Questions also asked how a robot could assist in a pandemic, to what extent robots will be able to replace human caregivers, and how a robot could assist in the participant's profession. Another question included in the analysis asked about the specific activities listed in the survey.

Interview questions not included in this analysis were introductory questions asking the participant about their profession, what care means to them, how the COVID-19 pandemic impacted them, and about their use of technology and robots. Other interview questions asked participants what guidelines they would set for robots in healthcare, if they can imagine situations in which a robot made its user uncomfortable, and how an assistive robot would change a patient's social life. The final portion of the interview consisted of three specific scenarios that consisted of ethical dilemmas that may arise if a patient had a live-in assistive robot. Participants then evaluated how they would deal with each dilemma and how they would prevent it from arising in the first place.

\subsection{Participants}

A total of 32 individuals were interviewed for this study. This included 6 doctors, 1 dentist, 1 medical student, 2 public health professors, 10 social workers, and 12 chaplains. Participants were recruited via a convenience sampling method, primarily being contacted through emails publicly listed on online university directories.

\subsubsection{Physicians, and other Healthcare Workers}
Numerous professors were emailed from the directory of affiliates of a medical school of interest, and a link to the survey was sent out in the medical school's weekly newsletter.  As mentioned, interviewed respondents included 6 doctors, 1 dentist, 1 medical student, and 2 public health professors. All of the interviewed physicians in addition to the dentist also worked either as educators at the medical school or as administrative personnel. The inclusion of public health professors was particularly interesting as public health focuses on a much broader level of care than a physician, social worker, or chaplain would. This perspective is of interest when designing robots to be integrated into the healthcare system as these participants have unique perspectives on the equity, access, and wider health impacts of a medical intervention. 

\subsubsection{Social Workers}
Ten social workers participated in the interview. This group was the most geographically diverse as participants were recruited from five schools in four different US states. Similar to the recruitment of some participants in the other two groups, social workers were identified through the websites of various schools of social work. Thus, most of the social workers were also professors. 

Social workers were included as interview participants due to their involvement in emotional and behavioral care. This type of care can be seen as distinct from the care provided by the doctors and other healthcare professionals who participated in the interview.

\subsubsection{Chaplains}

Many caregiving ecosystems, such as hospitals, provide access to spiritual care providers. These chaplains can offer advice to patients, their families, and other healthcare professionals on how to approach issues of life, death, quality of life, emotional  well-being, interactions with vulnerable fellow-others and other issues of compassionate care \cite{lefevor2021religiousness}. When considering the introduction of AI-enabled robots in the caregiving ecosystem, it is important to understand the norms that inform how chaplains provide care; however, rather than being authoritative, this exercise will allow roboticists to acknowledge the landscape that robots might enter. Even if not directly providing the type of care offered by chaplains, AI-enabled robots that undermine the chaplains' norms of caregiving might still threaten trust and increase distress in their limited assistive capacities.

Chaplains were identified by visiting the websites of chaplaincy departments of local hospitals and universities. Identified chaplains were then emailed and asked to participate in the survey and interview. 

All of the chaplains who were interviewed were from the same metropolitan area. University chaplains from three universities with major training programs in healthcare professions were interviewed. Hospital chaplains from one hospital were interviewed. One of the participants did not officially hold the title of chaplain but was instead a chief healthcare ethicist overseeing hospitals affiliated with a religious tradition.

Chaplains from a variety of traditions were invited to participate in our study: Africana, Bahai'i, Buddhist, Catholic, Church of Jesus Christ of Latter-Day Saints, Cru Agape, Lutheran, Episcopal, Evangelical, Greek Orthodox, Hindu, Humanist, Islamic, Jewish, Mar Thoma, Methodist, Sikh, and Unitarian Universalist. Participants representing the following traditions responded to our invitation and were interviewed: Catholic, Church of Jesus Christ of Latter-Day Saints, Greek Orthodox, Hindu, Humanist, Islamic, Jewish, Lutheran, and Methodist.

\subsection{Analytical approach}

We used an inductive and exploratory approach for identifying themes in the participants' answers.
Because we were interested in the participants' ideas and not the particularities of their speech, we have removed disfluencies and false starts from the quotes for easy readability. Each direct quote is followed by a randomly assigned participant number -- readers who are interested in following a specific person's thoughts can track quotes belonging to the person throughout the paper by using this number as a reference.

\section{Thoughts and attitudes about robots}

To ground ourselves in our analyses, we wanted to have an understanding of our participants' mental models of robots. This includes questions such as what do they first think of when they think of robots, what information sources do they draw from to form these mental models, what societal role(s) do they picture robots filling, and what concerns do they bring up about robots? To answer these questions, we looked at both the results from the pre-interview survey, as well as information peppered through the entirety of the interview itself.

We know from previous studies that people's mental models of robots are very fragile \cite{chita2021can}. Superficial factors such as a robot's appearance \cite{kwak2014impact,goetz2003matching,haring2018ffab,kwon2016human}, name \cite{law2021interplay,chita2019gender}, or movement patterns \cite{masuda2010motion} can have large-scale effects on people's expectations about the robot's social and emotional capabilities. Further, these factors affect the expectations not only of robot with which they are observing or interacting, but of all robots in general \cite{chita2021can}. Conversely, people's prior expectations about robots' appearances seem to depend on the application domain, even though people's visualizations of robots have many common features \cite{phillips2017does}. Additionally, outside factors such as media \cite{horstmann2019great,banks2020optimus} and culture \cite{bartneck2007influence} can shape people's mental models of robots as well. Given all of these notable influences on mental models of robots, we purposefully referred to robots in vague terms in our interview to not inadvertently prime participants. To get a summary of each participant's mental model, we asked them ``When you think of a robot what do you think of? Describe a robot to me.''

Some participants' answers to this question were inspired by science fiction, which has been shown to be a salient information source for novices' understandings of robots \cite{avin2019exploring}. Expectations about what a robot is capable of often mirror what has been shown in science fiction \cite{kriz2010fictional}. Participants who went down this path generally imagined a humanoid robot that can assist people with their daily lives similar to ``C3PO''(2199) or ``Rosie from the Jetsons''(3259). More general descriptions included a ``metal, humanoid-looking science fiction version''(1392), or something with ``a strange voice that moves awkwardly, doesn't show emotions, but can help out with different tasks''(7739). One participant explained ``I tend to think of a thing with eyes and ears and a nose and arms... I think that's because of popular depictions of robots in movies... if you grew up as I did in the 1950s, [that's your] view about what a robot is. I think it's been only in later years that we've come to appreciate that the term is a much more encompassing term''(1638). 

Other participants grounded their ideas of robots in ones that they had actually encountered, or knew existed, in the real world. The most frequent type of robot that these participants mentioned was a vacuum cleaning robot. For example, one chaplain mentioned ``I just bought my 85 year old father a [...] robotic vacuum cleaner''(7294). Other participants mentioned robotic toys, robots working in retail, or promotional videos they had seen on the internet of real robots. Another participant gave an anecdote of a machine that they had encountered: ``I went to the store yesterday -- the machine that took my bottles, you just fed the bottle in and scanned it and crushed it and packaged it on the other end''(9384). These two examples highlight how drawing inspiration from current robots leads to imagining more functional, less humanoid robots than those that are drawn from science fiction.

Some participants forewent embodiment entirely and described disembodied AI systems as their conception of a ``robot''. When asked how they can see a robot performing spiritual care, one chaplain talked about ``a device [providing] access to a vast amount of spiritual resources- prayers, hymns, meditations'' (9743). One social worker noted that ``a lot of jobs in social work are computing, they're not physical. So, you know, the idea of case management, being able to use artificial intelligence and machine learning to be able to help identify resources for people and hook them up with resources and make their appointments''(9528). Other participants imagined monitoring or emergency detection devices, with one even explicitly stated that a robot would not be necessary for this task: ``a robot could check on you, but I don't know that you need a full-blown robot to make sure that someone is up and about, because I know there's apps for your phone. I was looking at this for my parents- if you fall down and stop moving, the phone will call someone''(7989). These ideas highlight that there may be places where having a physically embodied robot could in fact be overkill.

Finally, some participants defined functional aspects of a robot in response to the question ``When you think of a robot what do you think of? Describe a robot to me''. One chaplain highlighted the lack of spirituality that a robot would have, saying ``anything that has an interface, that is algorithmic, as opposed to human which, I would say, our interface is spiritual instead of algorithmic''(3756). Another participant offered a different approach: ``it's really any mechanical device [or] technological device that can perform functions that ordinarily you'd have to do in a much more clumsy way''(5292). Many other participants mentioned general robot abilities, with almost all participants imagining that a robot that would be ``able to walk [and] talk''(2199). One participant explained that a robot would ``physically be able to move around whatever space, and it would communicate back to me based on what I said''(6801). One participant made a mechanical distinction, describing a ``robot that can walk and talk and function but is still a machine''(7392). Many participants also mentioned that a robot ``can help out with different tasks''(7739) and assist humans in their daily lives. Only one participant specifically stated that a robot ``can be controlled''(5279).  

The mental models that people have of robots, and the sources from which they build these models, can have different implications for what expectations and desires people have of robots. For example, the science fiction and media-based robots that people referenced, namely C3P0 and Rosie, both have high levels of social capability. They are also both humanoid, and the way that they complete tasks are often similar to how we might expect a human to, given the constraints and advantages offered by human morphology. Humanoid morphologies have been shown to create a greater expectation for social competency, and people assume that humanoid robots will be deployed in jobs that involve interacting with people \cite{zlotowski2020one}. Morphologies aside, media in particular has heightened people's expectations of robots in general \cite{sundar2016hollywood}. In contrast, the robots of today with which some participants had actually encountered are highly task-based, non-humanoid, and have no social skills (i.e., a vacuum cleaning robot or a bottle-crushing robot). Even robots that are sometimes found in retail which are meant to greet customers present only a facade of social competency \cite{wallach2008moral}. A person who expects to have Rosie-level interactions with one of those retail robots will either be disappointed or deceived by the social fluff that is presented as actual mental models of human interactants \cite{sharkey2011children}. 

Participants who brought up disembodied AI systems pinpointed tasks where having an embodied robot may in fact be unnecessary. Though there are advantages to embodiment, there are also drawbacks; for example, with current and near future technology, it would be difficult to create a single robot that can assist a disabled person with all of the physical tasks they need assistance with. Even if that robot were to be created, building and distributing these robots to all of the people who could possibly use them would be costly, both in time and money. Disembodied AI systems could be created to execute one or two small tasks with high proficiency with less worry about wasting resources by not capitalizing on fixing an entire problem at once. These AIs can be installed as phone apps or systems that are smaller and cheaper than physical robots, allowing them to be more accessible. A view of disembodied-AI in care, rather than fully embodied robots, could result in a greater proliferation of these technologies in the care ecosystem. Conversely, developing physically embodied robots, as many participants imagined them to be, would allow the robots to actively manipulate and change their environment. While the roles that these two differently embodied types of agents would play may have inherent differences, our participants often thought of them in tangent, opening a wide array of roles that robots could possibly play in care.

\section{Roles for robots}

Our main goal with this study was to uncover roles that robots could play in healthcare and caregiving as formulated by (health)care professionals. Through several questions and prompts we asked our participants to reflect on assistive roles that robots could play in different contexts and environments: people's daily lives in their own homes, in healthcare settings (e.g., hospitals, nursing homes etc.), and during an unusual health-relevant set of circumstances -- the COVID-19 pandemic. We also asked participants to specifically comment on whether and how they would see robots integrated as assistive devices in their own practice.
Participants mentioned a range of tasks that robots could assist with, spanning the spectrum from physical assistance (e.g., disinfecting, help with locomotion) to social assistance (e.g., nudges, companionship). For physical assistance, participants identified a range of tasks, from more impersonal and distal tasks (e.g., vacuuming the floor) to more proximal and personal tasks (e.g., help with bathing or grooming). The social assistance tasks mentioned covered the span of cognitive (e.g., reminders and nudges) to socio-emotional (e.g., companionship). The different perspectives given by the varied expertise of our group of participants was essential in covering the wide spectrum of tasks. We note, for example, that social workers were more likely to mention companionship and emotional assistance from robots than the other participant groups. The degree of disability that the robot was addressing also varied among participants. While some participants thought of robots that would simply improve well-being, others imagined robots assisting with severe impairments.

\subsection{Robots in daily lives}

When participants imagined robots integrated in the daily lives of people, they talked both about robots assisting with physical chores and robots performing socially assistive tasks. Table~1 shows a summary of all of the tasks that participants put forward. 

With regards to physical assistance, many participants suggested tasks that robots already commonly help with such as vacuuming, mowing the lawn, or cleaning a pool (see Table~1). Robots helping with such tasks have been widely publicized in media and advertising, so it is likely that participants drew from their knowledge of existing robots when answering this question. One example of such a task that was mentioned many times was transportation. Social workers were especially aware of this with many mentioning that ``transportation is a huge challenge for patients''(2199). Some participants even mentioned specific brands of self-driving cars.

When looking at physical daily life tasks that are not commonly thought of as robot roles, many participants would start listing off common chores, especially those that a disabled person may have challenges with. Many of the physical tasks that participants identified were given in this sort of way: participants would mention a challenge and some would give a short explanation of how a robot could help with the task.

Many participants however also imagined a robot that would perform care not through physical contact, but rather through cognitive and social support, which is what SARs are being designed for \cite{feil2005defining}. For example, participants imagined a robot that would serve to remind and nudge patients to do self-care activities. One social worker explained, ``I may see a client once a week or once every two weeks, and we may come up with sort of a plan to do certain things in between sessions, and a lot of times people get preoccupied and so they don't follow through... programming robots to sort of serve as [a reminder]''(4858). Another doctor went further to specify a specific condition. They explained that someone with a ``borderline IQ, to have a robot pal that would remind you how to make your bed, get up, be the `good morning it's time to get up, make your bed, brush your teeth, have your breakfast, oh did you have enough good breakfast'...some of them can't and some of them just don't, but I bet if they had a machine that did it, kind of a cute machine, it might be helpful''(2783). Other participatory-science studies surveying caregivers of people with dementia \cite{moharana2019robots} and workers with cognitive impairments \cite{williams2019aida} also expressed interest in robots assisting with nudges and reminders to help with diminished cognitive capacities due to health conditions.

However, some participants were unconvinced that robots would be able to help people with certain cognitive impairments, at least not if the robot was the sole assistance provider. One participant explained, ``I don't think a robot would be an option for care for someone living with memory impairment. That could be very scary, or for persons living with severe mental health disorders that may have hallucinations and delusions''(7392). Another participant explained that ``I couldn't see bathing, dressing, even feeding would be hard because most of the time people who need help feeding probably don't have the cognition to interact with [the robot]''(9103). These participants present an excellent point as many of the people who require assistance with daily activities may be too impaired to deal with a robot. Such a robot would need to be extremely advanced in order to perform enough activities to replace a caregiver. This clearly will not be available until farther into the future. As one participant explained, ``there's a natural progression -- we shouldn't move from the Roomba to The Jetsons''(4923). 

Another social worker mentioned that a robot would work in conjunction with a caregiver, ``not that it would replace a friendship or replace a family member looking in on a patient completely, but that it would just be that set of eyes and checking in, `hey it's eight o'clock in the morning, time to get up time to get take your shower and let's have breakfast...I'm gonna watch you take those meds and then oh it's 10 o'clock, time for you to log in and talk to your therapist, but before you do that here's some questions we need you to ask' all of that coming from the robot''(9384). Caregivers are often burdened with the repetitive nature of the tasks involved in monitoring someone with impaired cognitive capacities, and Moharana et al. \cite{moharana2019robots} have found support for a robotic companion that could ease the burden and resulting frustration from purely repetitive interactions.

Another social worker imagined a robot reminding someone ``to practice mindfulness or have you had breakfast today or it's time to do the laundry''(9103). One chaplain imagined a similar robot but for spiritual care: ``AI agent saying to you `let me give you your score card for the week. Here's where you are spiritually, here's how much you talked about yourself versus talking about other people in your conversations, here's how much time you spent in your apps that were teaching you principles that are of a more elevating and spiritual nature versus here's the amount of time you spent watching [TV]'''(0191). Using robots for enhancing well-being by supporting mindfulness and spiritual practices has been explored and implemented to a lesser extent in HRI, but a participatory design study investigating the use of virtual agents delivering health promotion interventions suggest that people respond very positively to agents that combine health content with personally and culturally relevant  messages, such as religious and scriptural references \cite{o2020community}. The participants were members of predominantly Black church communities. Additionally, the idea of providing users with well-being indicators (e.g., in the form of time spent focusing on self vs. others, or time spent consuming elevating content) has a parallel in the exploration of a different robotic application: that of providing feedback about inter-personal interactions (e.g., by keeping track of natural turns, speaking percentage, interactions etc.) \cite{tahir2014perception}. This shows that models of robots functioning as trackers and providers of feedback, and the robotic capabilities associated with this usage already exist and can be applied to health and well-being interventions. The idea of accurately tracking and offering people more insight into behaviors associated with well-being that are not typically monitored for one's health profile is a powerful one, and could be supported by assistive robots. For example, there is no systematic tracking of hopefulness or activities that enhance it even though it's been shown that changes in hopefulness has important effects on health \cite{long2020role}. By robots taking on this role, they could complement healthcare professionals and users could benefit from a more holistic approach to health and well-being.

Of all of the non-physical tasks mentioned by participants, these types of reminder and lifestyle robots were mentioned the most frequently. These tasks are more algorithmic than physical, so it is important to ask whether an embodied and likely more expensive robot is necessary. While self-care reminder apps already exist, an embodied robot may be more user-friendly with people who have trouble using technology as it may not require as much of a learning curve \cite{huber2016ethical}. Additionally, studies have shown that robots can exert social influence on people through their embodiment \cite{hayashi2019using} which might increase the likelihood that people will comply more to suggestions coming from a robot than a disemobided app. Such a robot has the potential to improve health and increase independence. The effectiveness of reminders in increasing the independence of patients with neurological diseases has been shown. One study found that when given prompts and positive reinforcement while eating unassisted, dementia patients improved their ability to eat independently \cite{coyne1997improving}.

In addition to imaginings of roles and tasks for robots, a few additional themes emerging from the interviews are worth pointing out: (1) the mental models caregiver have of {\em how} robots engage in these tasks, (2) the degrees of disability they imagined robotic interventions assisting with, and (3) independence as a goal of the robotic interventions. These themes emerged by chance in conversation (we did not intentionally probe for them), so only a small subset of participants contributed to these ideas; however, they can constitute good starting points for further research.

Some participants imagined a robot performing an activity in a way that would be very similar to a human. For example, when asked how a robot could help with shopping, one chaplain explained that the robot ``would have a map with all the aisles, and they would know where things are located and they would take the barcode of what has been put on a list into their system and go find it in the aisle and then pick it out of the aisle and drop it into after scanning into a bag and they would pay''(3756). However, some other participants  imagined an activity being done in a completely new way. In response to the same question, a doctor explained that a robot would ``put it all in a box and pick it up on a drone and deliver it to you''(5292). Participants did not comment directly on the implications of robots following (or not) a human model for executing tasks, but research in HRI is starting to pay attention to the way in which human-robot interactions are impacted by the way tasks are performed by the robot \cite{williams2014robot}. The shopping task described by the participants is impersonal and distal enough that one might not have any preferences for how it is performed, but for other more personal and intimate tasks, such as grooming and bathing people might have more clear expectations and preferences for how they should be done. Being mindful of what different stakeholders in the healthcare and caregiving ecosystem have in mind for how different tasks are performed is important input into the design of the robots, because it may have implications for robot acceptance.
 
The participants imagined robots assisting with various degrees of disability. The open-endedness of the question allowed participants to not only imagine {\em how} a robot would help, but also {\em what type of disability} the patient has. As one participant explained, a patient could be ``physically disabled'', but ``there's many different ways that you could be'' disabled (8201). It was common for participants to imagine a patient in a ``wheelchair''(7739). Or more generally someone with ``mobility issues'' or ``sight difficulties''(2199).
While Participants 4858 and 2783 focused on self-care tasks that would be beneficial to anyone's well-being, Participant 9384 focused on medical tasks such as complying with medication and treatment. This points to two ideas: (1) robots are perceived as covering a large part of, if not the entire spectrum of disability; and (2) the robot functionalities needed for interventions at a high degree of disability can find purpose at lower level as well, and maybe even for playing a role in prevention. This is in line with the idea of using {\em Universal Design} \cite{story2001principles} for developing robotic technologies. Universal Design principles advocate for designing objects to serve people across a spectrum of ability. As an example of Universal Design, it is worth taking another look, with disability levels in mind, at the nudge and reminder functionalities of robots that many participants brought up: they can be useful both for helping people maintain a higher-degree of emotional and spiritual well-being even outside of health concerns, and they can also be used to heavily support daily functioning for people with more severe cognitive impairments.

With respect to the goals of robotic assistance, participants consistently imagined robots that would help someone in being more independent than they would without a robot. Participant 9384 specifically mentioned that the person assisted would still have a human caregiver. Nonetheless, independence was mentioned as an important factor by our participants. Several participants used more general descriptions to explain that a robot could increase a patient's autonomy. One chaplain explained that having an assistive robot ``could be really liberating to have that 24-hour assistance''(6786). Another participant gave an anecdote: ``Both of my parents had Alzheimer's and my mother particularly sort of lived independently for a long time but she was declining and it was getting more and more problematic for her. We eventually had to put her in a nursing home. . . if a robot could have helped to kind of keep her in the place where she was living, I think that could have been maybe beneficial''(4858). Participants believed that allowing a patient to remain independent longer was beneficial. This is in agreement with both the goals set forth by assistive robotics \cite{van2020home,fiorini2021assistive} and in line with quality-of-life studies \cite{group1998world}. For example, one study  showed that elderly people prefer to age at home due to familiar surroundings and a sense of community \cite{wiles2012meaning}. Focusing on the person's independence is also paramount to avoiding enfeeblement, when the robot takes on too many of the daily living tasks resulting over time in a loss of capability and over-reliance on robots \cite{chita2021assistive}. 

Many of the roles envisioned for the robot as assistants in the daily lives of people, such as reminders, nudging to reinforce better health habits, encouraging self-care can be thought of as preventative care. Participants also saw roles for robots for more acute, health-critical circumstances which we will explore in the following sections.

\subsection{How robots can be a part of healthcare}

Many participants, especially social workers, saw the potential for robots assisting with therapy. Robot use for structured therapeutic protocols is an active area of research in HRI \cite{chita2021assistive}. One doctor explained that robots ``would be wonderful potential in terms of people with social anxieties, people with other sort of interpersonal disabilities like autism spectrum disorder'' and that they can be good for ``practicing conversations''(9528). A social worker had the same idea: ``robots would probably be easier for some people than talking with others, but as a means of developing social skills. A psychology tool''(1392). Another social worker mentioned ``exposure therapy...for people who have different types of maybe social anxieties [to] give them a non-threatening experience to practice with''(2199). Another participant imagined a robot providing feedback, explaining that the robot could ``reinforce behaviors, positive behaviors to model interactions... I think it could be programmed to interact with people who have different difficulty interacting, and even help them modulate their emotions''(9103). Again, many participants envisioned socially assistive robots, which indicates that participants felt that robots could assist people not just for physical health but also social and emotional health.

Using robots to train interaction in neuro-divergent individuals has been studied in HRI. For example, one study shows that autistic children who were assigned to interact with a robot spoke more than those interacting with a human or a computer game \cite{kim2013social}. However, while researchers in this study suggested that robots could potentially be used to reinforce behaviors, they were not able to determine whether robot interaction was beneficial to change behavior. Another study showed that an adaptive and personalized online treatment for social anxiety was effective \cite{helgadottir2014standalone}. However, this study is slightly different from what is being proposed by the interviewees. It involved pre-recorded sessions from clinical psychologists and functioned similarly to an adaptive online course. Further research is necessary to develop and study a treatment for social anxiety where an AI talks directly with the patient to parallel real-life social situations. Also, it should be evaluated whether embodiment of such an AI could add additional benefits.

One doctor explained that a robot would be effective for applied behavior analysis (ABA) which is a type of therapy that involves reinforcement to eliminate undesirable behavior in autism spectrum disorders. The doctor explained that a robot could be helpful in ABA because it is ``grueling work: it's repetitive [and] very costly.'' They explained that this treatment is only available to children who have ``families that can afford to send their kid to this school''(2783). Lowering healthcare costs and therefore expanding access to treatments is considered to be one of the key benefits of integrating AI and robots into healthcare \cite{wirtz2019cost}. Some participants, however, disagreed with the idea that robots could lower healthcare costs. One participant explained that ``the big limitation, like so much of this stuff, is going to be costly. People... are not going to pay 10-, 20-, 30,000 for a robot''(4128). In addition to the cost of creating a such a complex robot, mass production to lower costs may not be feasible as there are countless different disabilities each of which would require different robot features. One participant explained, ``the robot would just have to be tailored to the particular need of the person''(5501). 

In addition to emotional and social support for people, healthcare workers imagined many algorithmic tasks being taken over by robots. Several social workers specifically mentioned discharging a patient, and the tasks involved in releasing a patient from a clinical setting. One explained that ``a social worker in a health care setting traditionally does discharge, so they could do all the discharge with a robot''(6801). Another explained that ``some of the significant barriers we have when patients discharge from a hospital and go home is just that- that ability for them to get to appointments and because of either financial restraints around transportation or physical restraints around mobility and that could all be alleviated and you could have a social worker working wherever, sort of monitoring multiple patients through robots''(9384). Another social worker spoke more generally, explaining that ``a lot of jobs in social work are computing. They're not physical, so the idea of case management, being able to use artificial intelligence and machine learning to be able to help identify resources for people and hook them up with resources and make their appointments''(9528). Another participant mentioned ``taking patient histories and the data mining ''(5211) as a potential application of robots.

The HRI literature has suggested that robots would take over dull, dirty, and dangerous tasks \cite{fishel2020tactile}. Many of the healthcare professionals interviewed imagined robots capable of doing tasks that can be grouped into these categories. Several participants thought that robots were capable of performing dull, repetitive tasks, perhaps even surpassing human ability at performing them. One explained, ``I think that they would be able to substitute at the level of repetitive tasks, mundane tasks''(1392). One doctor even specified that robots ``probably do a better job than some caregivers, at least in terms of a sort of goal-oriented task of completing something accurately and in a repeated fashion without getting tired''(6889). Many participants believed that robots could be a huge asset in healthcare by taking over basic jobs and freeing up professionals to do work that is more meaningful to them and to the patient. For example, one participant said ``my mother had health care aides come in to to kind of help them clean the apartment, maybe help her take a shower, that sort of thing. And so I could see a robot doing that, and then allowing health care aides or aids people to come in and give that companionship. So they could focus on that while the robots could do the more the household chores''(4858). One public health professor explained the impact of freeing up a doctor's time: ``when you go to the doctor's, there's a lot of routine questions that you get asked- they take a history and they do a physical and whatever else. There's no reason the robot couldn't do that, and in fact I think a robot could do a much more thorough job because why? Because too many physicians in practice now are overworked''(1638). The effects of having an assistant to take over mundane and often unfulfilling tasks in healthcare have been studied. One study found that overall physician satisfaction increased when a scribe took over the routine and time-consuming task of recording information in electronic health records\cite{gidwani2017impact}. In addition to talking about dull tasks, one participant imagined a robot taking over some of the dirty jobs in healthcare by ``wiping somebody's [butt]... the basic, mundane, or unpleasant aspects of what we do for each other could be done by a robot''(9743). In healthcare, dirty and dangerous tasks tend to overlap as exposure to disease is an inherent risk of caring for the sick. This is particularly relevant for the roles of robots in health crises such as pandemics. 

\subsection{Robots' role in COVID/other pandemics}

The COVID-19 pandemic brought the dangers of working in healthcare into the public eye. As for the use of robots in healthcare, a pandemic could potentially add another layer of utility and urgency to using robots rather than humans (at least for some tasks). For this study, interviews were conducted from September 2020 to June 2021, so participants would frequently mention the pandemic. Additionally, the healthcare worker groups were specifically asked ``Do you envision any specific roles for robots in healthcare especially during the COVID-19 pandemic?'' All of the possible benefits of using robots in a pandemic that participants mentioned were related to reducing human to human contact, with the greatest benefit being that a robot could not act as a host for the virus and it ``could be cleaned after each patient''(6889).

Many of the ideas that were mentioned when discussing the potential roles for robots in healthcare or in daily life were also mentioned in the context of the COVID-19 pandemic. Overall, participants imagined that robots would help patients ``avoid some of the risks'' involved with COVID-19 ``by potentially obtaining things that people need so they don't have to potentially travel to them''(2199). More specifically, another participant explained that robots could help ``deliver groceries ... deliver medication''(5279). Robots would ``do the food shopping or go out and pick up things. If they needed a haircut, cutting their hair''(4858). 

The most common type of assistance participants imagined robots doing during a pandemic can be thought of as logistical help. Several participants even imagined a robot that could ``administer the vaccine''(5292) or a robot that could ``automate a lot of the testing''(4128). Some participants imagined robots eliminating bottlenecks in COVID related production. Explaining that robots can ``produce the vaccine''(7739) or ``make masks, or at least facilitate making masks''(9528). At the time of these interviews, the news was filled with stories of vaccine and PPE shortages. With these tasks, participants may have mentioned a robot simply because they knew of the issue rather than because a robot would be more efficient or desirable than a human.

Many participants imagined a robot working directly in healthcare. Some tasks were similar to those in daily care. One participant spoke about the benefits of ``within the hospital or the institutional setting ... [getting] food to people''(9103). One doctor spoke more generally, ``you wouldn't be worrying about having to to protect the caregivers, you wouldn't have to worry about having to protect yourself from the caregivers...a level of care could be given on a routine basis''(1392). Participants quickly saw the benefits of using healthcare robots in the pandemic, and many wished such a technology had been available. One doctor explained that ``it would have been great to have robots able to do a whole bunch of stuff in the healthcare arena in the early days of Covid because it would have reduced infection and transmission among healthcare workers''(1132). Another doctor imagined an expanded form of a technology already being used during the pandemic. They explained, ``currently the telehealth visits are limited in that you are not able to examine the patient in real time, you have to depend on the audio interaction or possibly a video interaction, but are not able to examine the individual. Having a device that could perform an examination would be very useful if we were to go back through a pandemic such as we did''(6889). One challenge with such an examination robot would be the need for it to interact with each patient. This would require the patient to either already have an assistive robot capable of examining them or to have a robot delivered to a patient for each appointment. Another doctor imagined a robot in a hospital: ``if you imagine a robot that had great tactile feedback, great imaging, you could place personal protective equipment on others by a robot- you wouldn't need a second person to help them tie up one of the gown[s]''(9272). This idea would minimize the need for providers to interact with each other while not in PPE. Another participant imagined ``a technology that could do virus detection at home [and] sanitize certain items''(7392). While the participant mentioned using virus detection in the home setting, such a device could potentially be useful in many environments, including clinical ones. Due to the high contagiousness and airborne transmission, virus detection has been an issue throughout the pandemic. One company called Roboscientific was able to create a sensor specifically for COVID-19 and claims to have near perfect accuracy \cite{guest2021using}.

While robot roles envisioned for daily life promotion of well-being or in assisting with (non-pandemic related) health conditions more directly were focused on socially-assistive robots, only one participant mentioned social robots in response to the COVID-19 question. They mentioned a robot assisting with ``some sort of social interaction and broadcast games and have some dialogue with an individual about music or what have you connect them you know to other worlds maybe tours and things like that''(7392).

\begin{table*}[ht]
\caption{\label{tab:Comparison} A summary of the robot roles our participants discussed, as organized by type of task (note: some tasks may appear in multiple columns as they fit multiple task types).}
\begin{adjustbox}{width=\textwidth}
\begin{tabular}{p{6cm}|p{6cm}|p{6cm}}
    \toprule
    \textbf{Physical Tasks} & \textbf{Cognitive Tasks} & \textbf{Social/Emotional Tasks}\\
    \hline
    \hline
    Vacuuming/cleaning & Playing games & Interacting with children \\
    \hline
    Getting dressed & Reading assistance & Facilitating conversation \\
    \hline
    Preparing food & Waking you up & Companionship \\
    \hline
    Controlling household appliances & Making appointments & Telling someone to exercise, do a breathing exercise, meditate, etc. \\
    \hline
    Mobility assistance & Maintaining checkbooks & Telling a visually impaired person about their surroundings and where to look during a social interaction \\
    \hline
    Cooking & Voice transcription & Pet-like companion \\
    \hline
    Mowing the lawn & Data management & Giving elderly people something to care for \\
    \hline
    Carrying things & File-keeping & Giving virtual tours \\
    \hline
    Deliveries & Reading scriptures aloud & Giving classes \\
    \hline
    Shopping & Working with patients who were discharged from the hospital & Playing games \\
    \hline
    Bathing & Registering patients for healthcare & Making appointments \\
    \hline
    Retrieving items throughout the house & Case management & Tracking behavior and giving weekly report on spirituality \\
    \hline
    Fall/emergency detection & Taking patient history & Suggesting appropriate scripture or music during a crisis \\
    \hline
    Using the toilet & Medication administration or reminders & Gathering information on health related quality of life \\
    \hline
    Cleaning a pool & Recording symptoms of a disease & Repetitive therapies for autism spectrum disorders \\
    \hline
    During Muslim prayer: helping with prostration, bowing, and washing feet & & Exposure and/or desensitization therapy for phobias or anxieties \\
    \hline
    Taking and recording vitals & & Helping to avert a crisis such as an anger episode or panic attack \\
    \hline
    Surgery & & Pointing someone to resources during a crisis \\
    \hline
    Early detection of disease & & Avoiding infection by using telepresence to replace in-person appointments \\
    \hline
    Examination of nerve function & & Helping socially anxious people prepare for a return to work/school \\
    \hline
    Giving injections/vaccines & & Providing companionship and interaction \\
    \hline
    Advanced prosthetic limb & & \\
    \hline
    Drawing blood & & \\
    \hline
    Rolling patients over in bed to avoid bed sores & & \\
    \hline
    Covid-19 testing & & \\
    \hline
    Producing and distributing PPE & & \\
    \hline
    At-home virus detection & & \\
    \hline
    Sanitizing & & \\
    \hline
    Cutting hair & & \\
    \bottomrule
\end{tabular}
\end{adjustbox}
\end{table*}

\section{Robot's potential for humanistic care}

This research project was based on the premise that caregiving is multidimensional and engages and affects many different life aspects of both the person receiving care and the caregiver. Caregiving is concerned not just with physical aspects of functioning and well-being, but also with social-emotional and even existential dimensions of people's lives. It is this multidimensional approach to the concept of caregiving that motivated the selection of our participants spanning healthcare professions that address the wide-range of needs. Our view of caregiving is compatible with the concept of ``humanistic'' care which has been linked to assistive robotics by Coghlan \cite{coghlan2021robots}. This view of care imagines patients who are not just cared for physically but also emotionally and socially in seemingly empathetic and thoughtful ways. Coghlan imagines a type of care that feels human even if it is provided by a robot.

When people engage in caregiving towards one-another they usually do this in the context of a relationship, that often combines to varying degrees multiple different dimensions of care: physical, social-emotional and existential-spiritual.  As we have shown in previous sections, people imagine robots performing tasks that may fit into all of these care dimensions. In this section we explore the following themes: (1) the extent to which robots can approach human levels of caregiving; (2) the extent to which this is desirable; and (3) in as much as robots enter the care ecosystem the ways in which they can do that meaningfully and beneficially without disrupting other care relationships.

One of the most heavily debated ethical issues in using robots for care is whether artificial relationships are inherently inferior to human-human relationships or even social at all \cite{scheutz201113}. Our participants most often imagined disabled or elderly people as receivers of robotic care. These groups of people tend to experience higher levels of social isolation which ultimately results in worse health outcomes; a human caregiver can represent a form of social interaction for these people \cite{singer2018health}. Human caregivers often provide not just assistance with physical needs but also social interaction and empathy, features of ``humanistic care'' \cite{coghlan2021robots}. Thus, in a case where a robot is replacing or even just supplementing a human caregiver it is important to not just consider whether the robot is capable of performing its required tasks, but if it is able to provide the social and emotional aspect of a human caregiver.

Even though some participants were in favor of robots assisting with more social aspects of care such as therapy in a narrow way (when it involved repetitive, highly-scripted tasks), many were adamantly opposed to using robots/AI for counseling, especially in high-stakes scenarios when people would be faced with deeply existential problems (e.g., the death of a child). This idea was most often brought up by chaplains and social workers. A chaplain explained that a robot would not be able to offer counseling because of its lack of empathy and that ``the connectivity of humanity would be lost with something artificial''(4923). A social worker explained: ``I don't envision robots being able to have a genuine emotional connection with somebody in that way, especially after they've experienced a crisis or a traumatic event''(5279). A hospital chaplain gave an example of a specific crisis: ``it seems like the human capacity to receive robotic care, I would think that would run into some real limitations around highly critical human experiences like death of a child... would you want a robot to roll in and say the prayers for your child who's about to die?''(9743). Earlier in the interview, Participant 9743 explained that they are frequently called upon to give counseling to the grieving family of a recently deceased patient. Participant 9743's personal experience offers an explanation as to why social workers and chaplains were the groups most opposed to robot-delivered counseling.

Several other participants who were unconvinced of robots' ability to maintain a human-human level relationship focused their concerns on robots providing companionship. ``I definitely think you can provide companionship but they maybe just can't provide a complex relationship, a truly human complex relationship''(9528). Another participant said: ``I have some hesitation about that in using something that's non-human to keep people company, but that aside, I could see something like that being useful''(4858). Almost all of the participants who talked about companionship were supportive of robots engaging in some level of emotional interaction and support as long as it did not become the user's only interaction. One participant explained that ``any task other than something that's deeply emotional or would require compassion or empathy I could see a robot doing''(5279). It is important to note that no participants mentioned social or emotional concerns when asked about robots in a pandemic. A possible explanation is that given the urgency of a pandemic robots were viewed as instruments able to prevent loneliness rather than to enhance it. 

Some participants emphasized that using robots in care can be a matter of preference and choice: ``It could be liberating for one person and oppressive to depressing to another''(6786).  Some participants even saw more advantages to human-robot interactions than human-human interactions for certain narrow domains: ``When I think about social and psychosocial issues, yeah for sure I think there's a lot of ways that I would be fine and in some ways I'd prefer to connect with a machine; sometimes having to build the rapport with a person is challenging or there's barriers there. I think too of being objective and I think of one of the issues that social workers spend a lot of time thinking about, [which] is just bias in the work that we do, bias that I bring to my interactions with clients, and I think robots could be maybe- there would have to be some thinking around it but-  completely free of bias''(9384). This suggests that instead of worrying about the equivalence between human caregiving and robotic caregiving it might be more appropriate to think of how robots could fit in a caregiving ecosystem where both humans and robots provide care.

Indeed, some participants expressed concerned not about the human robot interaction {\em per se} and its shortcomings in terms of humanistic care, but more about the effects the presence of the robot would have on the other human-human interactions. One participant explained that having a robot caregiver, even if it was not specifically designed to have a relationship with the patient, would reduce human to human interaction: ``for many folks, the caregiving person is their human connection. That's very meaningful, and so I wouldn't want to take that away...but that said, I know there's folks who really need the support, however it comes''(6786). Even though this participant was hesitant about replacing a human, they believed that it may be a necessary evil as it would expand access to those who did not have it in the first place. The worry that robots might reduce human-human interactions has been highlighted by many robot ethicists as well \cite{sharkey2010crying,sharkey2012granny}. As a solution, participants favored setting a threshold for the level of emotional care that should be done by a robot. Often this threshold was ambiguous and participants believed that robots would not be able to ``replace human care'' but rather ``work collaboratively with human service providers''(7392). One participant noted: ``Not that it would replace a friendship or replace a family member looking in on a patient completely, but that it would just be that set of eyes and checking in" (9384). Another idea was using ``technology as a medium, alleviating loneliness to a certain extent like Zoom, Whatsapp'' but not using a robot as the actual ``subject itself [to] alleviate loneliness''(2543).

\section{Conclusions and implications for HRI}

It is encouraging to see many of the uses of robots that are being explored in HRI echoed in the participants' interviews. This suggests that robots and their assistance is unlikely to be rejected just because they are robots.  It is particularly noteworthy that healthcare professionals view robots as suitable not just for performing physical and menial tasks, but see them as providing cognitive, social and even emotional support, which is exactly what the class of socially assistive robots (SARs) are meant for \cite{feil2005defining}. It is also notable that participants talk fairly favorably about robots assisting with very proximal and personal tasks such as bathing. It remains to be seen whether receivers of care would be welcoming of such robotic interventions. 

The interviews conducted suggest that robots could be useful for all dimensions of care, from physical, to social-emotional and even, through performing narrow tasks, for supporting the spiritual-existential dimension of care. Participants did stress limitations in the robots' ability to provide comprehensive humanistic care; however, by complementing rather than replacing human caregivers, robots were generally seen as potentially useful rather than disruptive additions to the the healthcare ecosystem. 

Potential problems with the introduction of robots in the caregiving ecosystem were noted not just in relation to robots capabilities (or lack thereof), but also with how they might be used. The idea of humanistic care highlights the fact that the different dimensions of care are not neatly separated but rather intertwined, for example assistance with a physical task can also provide emotional comfort. A possible solution pointed out by participants was a mindful and limited use of robots in caregiving so as not to lead to a neglectful approach to deeper aspects of care that they cannot fulfill. Overall the results suggest that working on both improving robot capabilities as well as creating sound frameworks for the circumstances and extent to which robots should be used would ensure that they are a welcome and beneficial addition to the healthcare and caregiving ecosystem.

\begin{acknowledgements}

\end{acknowledgements}

\section{Conflict of interest}

The authors declare that they have no conflict of interest.

\bibliographystyle{plain}     

\bibliography{references}   

\begin{thebibliography}{10}

\bibitem{avin2019exploring}
Shahar Avin.
\newblock Exploring artificial intelligence futures.
\newblock {\em Journal of AI Humanities. Available at https://doi.
  org/10.17863/CAM}, 35812:1440, 2019.

\bibitem{banks2020optimus}
Jaime Banks.
\newblock Optimus primed: Media cultivation of robot mental models and social
  judgments.
\newblock {\em Frontiers in Robotics and AI}, 7:62, 2020.

\bibitem{bartneck2007influence}
Christoph Bartneck, Tomohiro Suzuki, Takayuki Kanda, and Tatsuya Nomura.
\newblock The influence of people’s culture and prior experiences with aibo
  on their attitude towards robots.
\newblock {\em Ai \& Society}, 21(1-2):217--230, 2007.

\bibitem{bedaf2015overview}
Sandra Bedaf, Gert~Jan Gelderblom, and Luc De~Witte.
\newblock Overview and categorization of robots supporting independent living
  of elderly people: what activities do they support and how far have they
  developed.
\newblock {\em Assistive Technology}, 27(2):88--100, 2015.

\bibitem{block2019softness}
Alexis~E Block and Katherine~J Kuchenbecker.
\newblock Softness, warmth, and responsiveness improve robot hugs.
\newblock {\em International Journal of Social Robotics}, 11(1):49--64, 2019.

\bibitem{bower1998cognitive}
Julienne~E Bower, Margaret~E Kemeny, Shelley~E Taylor, and John~L Fahey.
\newblock Cognitive processing, discovery of meaning, cd4 decline, and
  aids-related mortality among bereaved hiv-seropositive men.
\newblock {\em Journal of consulting and clinical psychology}, 66(6):979, 1998.

\bibitem{candela2020finding}
Maria~Luigia Candela, Michela Piredda, Anna Marchetti, Gabriella Facchinetti,
  Laura Iacorossi, Maria~Teresa Capuzzo, Daniela Mecugni, Laura Rasero, Maria
  Matarese, and Maria~Grazia De~Marinis.
\newblock Finding meaning in life: an exploration on the experiences with
  dependence on care of patients with advanced cancer and nurses caring for
  them.
\newblock {\em Supportive Care in Cancer}, 28(9):4493--4499, 2020.

\bibitem{chita2021can}
Meia Chita-Tegmark, Theresa Law, Nicholas Rabb, and Matthias Scheutz.
\newblock Can you trust your trust measure?
\newblock In {\em Proceedings of the 2021 ACM/IEEE International Conference on
  Human-Robot Interaction}, pages 92--100, 2021.

\bibitem{chita2019gender}
Meia Chita-Tegmark, Monika Lohani, and Matthias Scheutz.
\newblock Gender effects in perceptions of robots and humans with varying
  emotional intelligence.
\newblock In {\em 2019 14th ACM/IEEE International Conference on Human-Robot
  Interaction (HRI)}, pages 230--238. IEEE, 2019.

\bibitem{chita2021assistive}
Meia Chita-Tegmark and Matthias Scheutz.
\newblock Assistive robots for the social management of health: a framework for
  robot design and human--robot interaction research.
\newblock {\em International Journal of Social Robotics}, 13(2):197--217, 2021.

\bibitem{coghlan2021robots}
Simon Coghlan.
\newblock Robots and the possibility of humanistic care.
\newblock {\em International Journal of Social Robotics}, pages 1--14, 2021.

\bibitem{conte2020design}
Dean Conte, Spencer Leamy, and Tomonari Furukawa.
\newblock Design and map-based teleoperation of a robot for disinfection of
  covid-19 in complex indoor environments.
\newblock In {\em 2020 IEEE International Symposium on Safety, Security, and
  Rescue Robotics (SSRR)}, pages 276--282. IEEE, 2020.

\bibitem{coyne1997improving}
Mary~Louise Coyne and Lois Hoskins.
\newblock Improving eating behaviors in dementia using behavioral strategies.
\newblock {\em Clinical nursing research}, 6(3):275--290, 1997.

\bibitem{davies2000review}
Brian Davies.
\newblock A review of robotics in surgery.
\newblock {\em Proceedings of the Institution of Mechanical Engineers, Part H:
  Journal of Engineering in Medicine}, 214(1):129--140, 2000.

\bibitem{feil2005defining}
David Feil-Seifer and Maja~J Mataric.
\newblock Defining socially assistive robotics.
\newblock In {\em 9th International Conference on Rehabilitation Robotics,
  2005. ICORR 2005.}, pages 465--468. IEEE, 2005.

\bibitem{fiorini2019assistive}
Laura Fiorini, Marleen De~Mul, Isabelle Fabbricotti, Raffaele Limosani,
  Alessandra Vitanza, Grazia D’Onofrio, Michael Tsui, Daniele Sancarlo,
  Francesco Giuliani, Antonio Greco, et~al.
\newblock Assistive robots to improve the independent living of older persons:
  results from a needs study.
\newblock {\em Disability and Rehabilitation: Assistive Technology}, pages
  1--11, 2019.

\bibitem{fiorini2021assistive}
Laura Fiorini, Marleen De~Mul, Isabelle Fabbricotti, Raffaele Limosani,
  Alessandra Vitanza, Grazia D’Onofrio, Michael Tsui, Daniele Sancarlo,
  Francesco Giuliani, Antonio Greco, et~al.
\newblock Assistive robots to improve the independent living of older persons:
  results from a needs study.
\newblock {\em Disability and Rehabilitation: Assistive Technology},
  16(1):92--102, 2021.

\bibitem{fischinger2016hobbit}
David Fischinger, Peter Einramhof, Konstantinos Papoutsakis, Walter Wohlkinger,
  Peter Mayer, Paul Panek, Stefan Hofmann, Tobias Koertner, Astrid Weiss,
  Antonis Argyros, et~al.
\newblock Hobbit, a care robot supporting independent living at home: First
  prototype and lessons learned.
\newblock {\em Robotics and Autonomous Systems}, 75:60--78, 2016.

\bibitem{fishel2020tactile}
Jeremy~A Fishel, Toni Oliver, Michael Eichermueller, Giuseppe Barbieri, Ethan
  Fowler, Toivo Hartikainen, Luke Moss, and Rich Walker.
\newblock Tactile telerobots for dull, dirty, dangerous, and inaccessible
  tasks.
\newblock In {\em 2020 IEEE International conference on robotics and automation
  (ICRA)}, pages 11305--11310. IEEE, 2020.

\bibitem{gidwani2017impact}
Risha Gidwani, Cathina Nguyen, Alexis Kofoed, Catherine Carragee, Tracy Rydel,
  Ian Nelligan, Amelia Sattler, Megan Mahoney, and Steven Lin.
\newblock Impact of scribes on physician satisfaction, patient satisfaction,
  and charting efficiency: a randomized controlled trial.
\newblock {\em The Annals of Family Medicine}, 15(5):427--433, 2017.

\bibitem{goetz2003matching}
Jennifer Goetz, Sara Kiesler, and Aaron Powers.
\newblock Matching robot appearance and behavior to tasks to improve
  human-robot cooperation.
\newblock In {\em The 12th IEEE International Workshop on Robot and Human
  Interactive Communication, 2003. Proceedings. ROMAN 2003.}, pages 55--60.
  Ieee, 2003.

\bibitem{group1998world}
The~Whoqol Group.
\newblock The world health organization quality of life assessment (whoqol):
  development and general psychometric properties.
\newblock {\em Social science \& medicine}, 46(12):1569--1585, 1998.

\bibitem{guest2021using}
C~Guest, SY~Dewhirst, DJ~Allen, S~Aziz, O~Baerenbold, and J~Bradley.
\newblock Using trained dogs and organic semi-conducting sensors to identify
  asymptomatic and mild sars-cov-2 infections, 2021.

\bibitem{guettari2020uvc}
Moez Guettari, Ines Gharbi, and Samir Hamza.
\newblock Uvc disinfection robot.
\newblock {\em Environmental Science and Pollution Research}, pages 1--6, 2020.

\bibitem{haring2018ffab}
Kerstin~S Haring, Katsumi Watanabe, Mari Velonaki, Chad~C Tossell, and Victor
  Finomore.
\newblock Ffab—the form function attribution bias in human--robot
  interaction.
\newblock {\em IEEE Transactions on Cognitive and Developmental Systems},
  10(4):843--851, 2018.

\bibitem{hayashi2019using}
Yugo Hayashi, Kosuke Wakabayashi, Shigen Shimojyo, and Yukoh Kida.
\newblock Using decision support systems for juries in court: comparing the use
  of real and cg robots.
\newblock In {\em 2019 14th ACM/IEEE International Conference on Human-Robot
  Interaction (HRI)}, pages 556--557. IEEE, 2019.

\bibitem{helgadottir2014standalone}
Fj{\'o}la~D{\"o}gg Helgad{\'o}ttir, Ross~G Menzies, Mark Onslow, Ann Packman,
  and Sue O’Brian.
\newblock A standalone internet cognitive behavior therapy treatment for social
  anxiety in adults who stutter: Cbtpsych.
\newblock {\em Journal of Fluency Disorders}, 41:47--54, 2014.

\bibitem{hinds1988hopefulness}
Pamela~S Hinds and Janni Martin.
\newblock Hopefulness and the self-sustaining process in adolescents with
  cancer.
\newblock {\em Nursing Research}, 1988.

\bibitem{horstmann2019great}
Aike~C Horstmann and Nicole~C Kr{\"a}mer.
\newblock Great expectations? relation of previous experiences with social
  robots in real life or in the media and expectancies based on qualitative and
  quantitative assessment.
\newblock {\em Frontiers in psychology}, 10:939, 2019.

\bibitem{huber2016ethical}
Andreas Huber, Astrid Weiss, and Marjo Rauhala.
\newblock The ethical risk of attachment how to identify, investigate and
  predict potential ethical risks in the development of social companion
  robots.
\newblock In {\em 2016 11th ACM/IEEE International Conference on Human-Robot
  Interaction (HRI)}, pages 367--374. IEEE, 2016.

\bibitem{james2018artificial}
Jesin James, Catherine~Inez Watson, and Bruce MacDonald.
\newblock Artificial empathy in social robots: An analysis of emotions in
  speech.
\newblock In {\em 2018 27th IEEE International Symposium on Robot and Human
  Interactive Communication (RO-MAN)}, pages 632--637. IEEE, 2018.

\bibitem{jecker2020you}
Nancy~S Jecker.
\newblock You’ve got a friend in me: sociable robots for older adults in an
  age of global pandemics.
\newblock {\em Ethics and Information Technology}, pages 1--9, 2020.

\bibitem{kaplan1977social}
Berton~H Kaplan, John~C Cassel, and Susan Gore.
\newblock Social support and health.
\newblock {\em Medical care}, 15(5):47--58, 1977.

\bibitem{kim2013social}
Elizabeth~S Kim, Lauren~D Berkovits, Emily~P Bernier, Dan Leyzberg, Frederick
  Shic, Rhea Paul, and Brian Scassellati.
\newblock Social robots as embedded reinforcers of social behavior in children
  with autism.
\newblock {\em Journal of autism and developmental disorders},
  43(5):1038--1049, 2013.

\bibitem{kriz2010fictional}
Sarah Kriz, Toni~D Ferro, Pallavi Damera, and John~R Porter.
\newblock Fictional robots as a data source in hri research: Exploring the link
  between science fiction and interactional expectations.
\newblock In {\em 19th international symposium in robot and human interactive
  communication}, pages 458--463. IEEE, 2010.

\bibitem{kwak2014impact}
Sonya~S Kwak.
\newblock The impact of the robot appearance types on social interaction with a
  robot and service evaluation of a robot.
\newblock {\em Archives of Design Research}, 27(2):81--93, 2014.

\bibitem{kwon2016human}
Minae Kwon, Malte~F Jung, and Ross~A Knepper.
\newblock Human expectations of social robots.
\newblock In {\em 2016 11th ACM/IEEE International Conference on Human-Robot
  Interaction (HRI)}, pages 463--464. IEEE, 2016.

\bibitem{kyrarini2021survey}
Maria Kyrarini, Fotios Lygerakis, Akilesh Rajavenkatanarayanan, Christos
  Sevastopoulos, Harish~Ram Nambiappan, Kodur~Krishna Chaitanya, Ashwin~Ramesh
  Babu, Joanne Mathew, and Fillia Makedon.
\newblock A survey of robots in healthcare.
\newblock {\em Technologies}, 9(1):8, 2021.

\bibitem{law2021interplay}
Theresa Law, Meia Chita-Tegmark, and Matthias Scheutz.
\newblock The interplay between emotional intelligence, trust, and gender in
  human--robot interaction.
\newblock {\em International Journal of Social Robotics}, 13(2):297--309, 2021.

\bibitem{lefevor2021religiousness}
G~Tyler Lefevor, Jacqueline~Y Paiz, Hannah~E Milburn, Paige~E Sheffield, and
  Nathalie~A Tamez~Guerrero.
\newblock Religiousness and help seeking: individual, congregational, and
  clergy factors.
\newblock {\em Counselling Psychology Quarterly}, pages 1--21, 2021.

\bibitem{long2020role}
Katelyn~NG Long, Eric~S Kim, Ying Chen, Matthew~F Wilson, Everett~L
  Worthington~Jr, and Tyler~J VanderWeele.
\newblock The role of hope in subsequent health and well-being for older
  adults: An outcome-wide longitudinal approach.
\newblock {\em Global Epidemiology}, 2:100018, 2020.

\bibitem{maalouf2018robotics}
Noel Maalouf, Abbas Sidaoui, Imad~H Elhajj, and Daniel Asmar.
\newblock Robotics in nursing: a scoping review.
\newblock {\em Journal of Nursing Scholarship}, 50(6):590--600, 2018.

\bibitem{martinez2020socially}
Ester Martinez-Martin, Felix Escalona, and Miguel Cazorla.
\newblock Socially assistive robots for older adults and people with autism: An
  overview.
\newblock {\em Electronics}, 9(2):367, 2020.

\bibitem{mast2012user}
Marcus Mast, Michael Burmester, Katja Kruger, Sascha Fatikow, Georg Arbeiter,
  Birgit Graf, Gernot Kronreif, Lucia Pigini, David Facal, and Renxi Qiu.
\newblock User-centered design of a dynamic-autonomy remote interaction concept
  for manipulation-capable robots to assist elderly people in the home.
\newblock {\em Journal of Human-Robot Interaction}, 2012.

\bibitem{masuda2010motion}
Megumi Masuda and Shohei Kato.
\newblock Motion rendering system for emotion expression of human form robots
  based on laban movement analysis.
\newblock In {\em 19Th international symposium in robot and human interactive
  communication}, pages 324--329. IEEE, 2010.

\bibitem{moharana2019robots}
Sanika Moharana, Alejandro~E Panduro, Hee~Rin Lee, and Laurel~D Riek.
\newblock Robots for joy, robots for sorrow: community based robot design for
  dementia caregivers.
\newblock In {\em 2019 14th ACM/IEEE International Conference on Human-Robot
  Interaction (HRI)}, pages 458--467. IEEE, 2019.

\bibitem{mok2010health}
Esther Mok, Ka-po Lau, Wai-man Lam, Lai-ngor Chan, Jeffrey Ng, and Kin-sang
  Chan.
\newblock Health-care professionals' perspective on hope in the palliative care
  setting.
\newblock {\em Journal of palliative medicine}, 13(7):877--883, 2010.

\bibitem{murphy2020applications}
Robin~R Murphy, Vignesh Babu~Manjunath Gandudi, and Justin Adams.
\newblock Applications of robots for covid-19 response.
\newblock {\em arXiv preprint arXiv:2008.06976}, 2020.

\bibitem{o2020community}
Teresa~K O'Leary, Elizabeth Stowell, Everlyne Kimani, Dhaval Parmar, Stefan
  Olafsson, Jessica Hoffman, Andrea~G Parker, Michael~K Paasche-Orlow, and
  Timothy Bickmore.
\newblock Community-based cultural tailoring of virtual agents.
\newblock In {\em Proceedings of the 20th ACM International Conference on
  Intelligent Virtual Agents}, pages 1--8, 2020.

\bibitem{parks2010lifting}
Jennifer~A Parks.
\newblock Lifting the burden of women's care work: should robots replace the
  “human touch”?
\newblock {\em Hypatia}, 25(1):100--120, 2010.

\bibitem{phillips2017does}
Elizabeth Phillips, Daniel Ullman, Maartje~MA de~Graaf, and Bertram~F Malle.
\newblock What does a robot look like?: A multi-site examination of user
  expectations about robot appearance.
\newblock In {\em Proceedings of the human factors and ergonomics society
  annual meeting}, volume~61, pages 1215--1219. SAGE Publications Sage CA: Los
  Angeles, CA, 2017.

\bibitem{rabbitt2015integrating}
Sarah~M Rabbitt, Alan~E Kazdin, and Brian Scassellati.
\newblock Integrating socially assistive robotics into mental healthcare
  interventions: Applications and recommendations for expanded use.
\newblock {\em Clinical psychology review}, 35:35--46, 2015.

\bibitem{ray2008people}
C{\'e}line Ray, Francesco Mondada, and Roland Siegwart.
\newblock What do people expect from robots?
\newblock In {\em 2008 IEEE/RSJ International Conference on Intelligent Robots
  and Systems}, pages 3816--3821. IEEE, 2008.

\bibitem{vsabanovic2013paro}
Selma {\v{S}}abanovi{\'c}, Casey~C Bennett, Wan-Ling Chang, and Lesa Huber.
\newblock Paro robot affects diverse interaction modalities in group sensory
  therapy for older adults with dementia.
\newblock In {\em 2013 IEEE 13th international conference on rehabilitation
  robotics (ICORR)}, pages 1--6. IEEE, 2013.

\bibitem{scassellati2020potential}
Brian Scassellati and Marynel V{\'a}zquez.
\newblock The potential of socially assistive robots during infectious disease
  outbreaks.
\newblock {\em Science Robotics}, 5(44), 2020.

\bibitem{scheutz201113}
Matthias Scheutz.
\newblock The inherent dangers of unidirectional emotional bonds between humans
  and social robots.
\newblock {\em Robot ethics: The ethical and social implications of robotics},
  page 205, 2011.

\bibitem{schwarzer1991social}
Ralf Schwarzer and Anja Leppin.
\newblock Social support and health: A theoretical and empirical overview.
\newblock {\em Journal of social and personal relationships}, 8(1):99--127,
  1991.

\bibitem{sharkey2011children}
Amanda Sharkey and Noel Sharkey.
\newblock Children, the elderly, and interactive robots.
\newblock {\em IEEE Robotics \& Automation Magazine}, 18(1):32--38, 2011.

\bibitem{sharkey2012granny}
Amanda Sharkey and Noel Sharkey.
\newblock Granny and the robots: ethical issues in robot care for the elderly.
\newblock {\em Ethics and information technology}, 14(1):27--40, 2012.

\bibitem{sharkey2010crying}
Noel Sharkey and Amanda Sharkey.
\newblock The crying shame of robot nannies: an ethical appraisal.
\newblock {\em Interaction Studies}, 11(2):161--190, 2010.

\bibitem{singer2018health}
Clifford Singer.
\newblock Health effects of social isolation and loneliness.
\newblock {\em J. Aging Life Care}, 28:4--8, 2018.

\bibitem{sparrow2006hands}
Robert Sparrow and Linda Sparrow.
\newblock In the hands of machines? the future of aged care.
\newblock {\em Minds and Machines}, 16(2):141--161, 2006.

\bibitem{story2001principles}
Molly~Follette Story.
\newblock Principles of universal design.
\newblock {\em Universal design handbook}, 2001.

\bibitem{sundar2016hollywood}
S~Shyam Sundar, T~Franklin Waddell, and Eun~Hwa Jung.
\newblock The hollywood robot syndrome media effects on older adults' attitudes
  toward robots and adoption intentions.
\newblock In {\em 2016 11th ACM/IEEE International Conference on Human-Robot
  Interaction (HRI)}, pages 343--350. IEEE, 2016.

\bibitem{tahir2014perception}
Yasir Tahir, Umer Rasheed, Shoko Dauwels, and Justin Dauwels.
\newblock Perception of humanoid social mediator in two-person dialogs.
\newblock In {\em Proceedings of the 2014 ACM/IEEE international conference on
  Human-robot interaction}, pages 300--301, 2014.

\bibitem{trivedi2014characteristics}
Ranak Trivedi, Kristine Beaver, Erin~D Bouldin, Evercita Eugenio, Steven~B
  Zeliadt, Karin Nelson, Ann-Marie Rosland, Jackie~G Szarka, and John~D Piette.
\newblock Characteristics and well-being of informal caregivers: Results from a
  nationally-representative us survey.
\newblock {\em Chronic Illness}, 10(3):167--179, 2014.

\bibitem{van2020home}
Ryan Van~Patten, Amber~V Keller, Jacqueline~E Maye, Dilip~V Jeste, Colin Depp,
  Laurel~D Riek, and Elizabeth~W Twamley.
\newblock Home-based cognitively assistive robots: maximizing cognitive
  functioning and maintaining independence in older adults without dementia.
\newblock {\em Clinical Interventions in Aging}, 15:1129, 2020.

\bibitem{van2013designing}
Aimee Van~Wynsberghe.
\newblock Designing robots for care: Care centered value-sensitive design.
\newblock {\em Science and engineering ethics}, 19(2):407--433, 2013.

\bibitem{walden2015mental}
Justin Walden, Eun~Hwa Jung, S~Shyam Sundar, and Ariel~Celeste Johnson.
\newblock Mental models of robots among senior citizens: An interview study of
  interaction expectations and design implications.
\newblock {\em Interaction Studies}, 16(1):68--88, 2015.

\bibitem{wallach2008moral}
Wendell Wallach and Colin Allen.
\newblock {\em Moral machines: Teaching robots right from wrong}.
\newblock Oxford University Press, 2008.

\bibitem{wiles2012meaning}
Janine~L Wiles, Annette Leibing, Nancy Guberman, Jeanne Reeve, and Ruth~ES
  Allen.
\newblock The meaning of “aging in place” to older people.
\newblock {\em The gerontologist}, 52(3):357--366, 2012.

\bibitem{williams2019aida}
Andrew~B Williams, Rosa~M Williams, Ronald~E Moore, and Matthias McFarlane.
\newblock Aida: a social co-robot to uplift workers with intellectual and
  developmental disabilities.
\newblock In {\em 2019 14th ACM/IEEE International Conference on Human-Robot
  Interaction (HRI)}, pages 584--585. IEEE, 2019.

\bibitem{williams2014robot}
Tom Williams, Priscilla Briggs, Nathaniel Pelz, and Matthias Scheutz.
\newblock Is robot telepathy acceptable? investigating effects of nonverbal
  robot-robot communication on human-robot interaction.
\newblock In {\em The 23rd IEEE international symposium on robot and human
  interactive communication}, pages 886--891. IEEE, 2014.

\bibitem{wirtz2019cost}
Jochen Wirtz.
\newblock Cost-effective service excellence in healthcare.
\newblock {\em AMS Review}, 9(1):98--104, 2019.

\bibitem{zlotowski2020one}
Jakub Z{\l}otowski, Ashraf Khalil, and Salam Abdallah.
\newblock One robot doesn’t fit all: aligning social robot appearance and job
  suitability from a middle eastern perspective.
\newblock {\em AI \& SOCIETY}, 35(2):485--500, 2020.

\end{thebibliography}

\end{document}